\newcommand{\mytitle}{180-degree Outpainting from a Single Image}
\newcommand{\mykeywords}{large field of view, spatially augmented reality, immersion, extrafoveal video} % augmented video, human vision, video extrapolation

\def\myversion{100} % final
\def\myorcid{0000-0001-9730-5262} % final

\documentclass[journal]{IEEEtran}

\def\mypath{./\jobname}
\def\mygraphicspath{\mypath/image}

\def\mymain{main.tex}

\def\mymain{main_body.tex}

\def\mygraphicspath{image}
\usepackage{enumitem}
\usepackage{xcolor}
\usepackage{amssymb}
\usepackage{amsmath}
\usepackage{lipsum}
\usepackage{graphicx} \graphicspath{{\mygraphicspath/}}

\usepackage{flushend} 

\usepackage{mdframed}
\usepackage{multirow}
\usepackage{booktabs}
\usepackage{soul}
\usepackage{xcolor}
\usepackage{gensymb}
\usepackage{amsmath}
\usepackage{amssymb}
\usepackage{hyperref}

\ifdefined\myversion %%%%%%%%%%%%%%%%%%%
\ifdefined\mode 
\else
  \def\mode{release}
  \def\myversion{100} % final
\fi

\ifdefined\release 
  \csname\mode\endcsname
\else
  
\fi

\ifnum0\myversion=100
  \usepackage[final]{changes}
  \newcommand{\note}[2][]{}
\else
  \usepackage[markup=todo]{changes}
  \newcommand{\note}[2][]{\added[#1,remark={#2}]{}}
  \usepackage{todonotes}
  \makeatletter
  \setremarkmarkup{\todo[color=Changes@Color#1!20,size=\scriptsize]{#1: #2}}
  \makeatother
\fi

\ifdefined\IEEEPARstart
\else
  \newcommand{\IEEEPARstart}[2]{#1#2}
\fi
\ifdefined\appendices
\else
  \newcommand{\appendices}{\section*{Appendices}}
\fi
\ifdefined\IEEEraisesectionheading
\else
  \newcommand{\IEEEraisesectionheading}[1]{#1}
\fi

\newcommand{\forconf}[1]{
\@ifundefined{\IEEEkeywords}{#1}
} \fi
\usepackage{bm}  % 斜体加粗体
\usepackage{xspace}

\makeatletter
\DeclareRobustCommand\onedot{\futurelet\@let@token\@onedot}
\def\@onedot{\ifx\@let@token.\else.\null\fi\xspace}

\makeatother

\ifdefined\myorcid 
  \usepackage{scalerel}
  \usepackage{tikz}
  \usetikzlibrary{svg.path}
  \definecolor{orcidlogocol}{HTML}{A6CE39}
  \tikzset{
    orcidlogo/.pic={
      \fill[orcidlogocol] svg{M256,128c0,70.7-57.3,128-128,128C57.3,256,0,198.7,0,128C0,57.3,57.3,0,128,0C198.7,0,256,57.3,256,128z};
      \fill[white] svg{M86.3,186.2H70.9V79.1h15.4v48.4V186.2z}
                   svg{M108.9,79.1h41.6c39.6,0,57,28.3,57,53.6c0,27.5-21.5,53.6-56.8,53.6h-41.8V79.1z M124.3,172.4h24.5c34.9,0,42.9-26.5,42.9-39.7c0-21.5-13.7-39.7-43.7-39.7h-23.7V172.4z}
                   svg{M88.7,56.8c0,5.5-4.5,10.1-10.1,10.1c-5.6,0-10.1-4.6-10.1-10.1c0-5.6,4.5-10.1,10.1-10.1C84.2,46.7,88.7,51.3,88.7,56.8z};
    }
  }

  \newcommand\orcidicon[1]{\href{https://orcid.org/#1}{\mbox{\scalerel*{
  \begin{tikzpicture}[yscale=-1,transform shape]
  \pic{orcidlogo};
  \end{tikzpicture}
  }{|}}}}
\fi
\ifdefined\majorrevision 
  
  \newcommand\highlight[2]{%
  \bgroup
  \hskip0pt\color{#1}%
  #2%
  \egroup
}

\else

\fi

\makeatletter
\def\input@path{{../}{./}{\mygraphicspath/}{\mypath/}{\jobname/}}
\makeatother

\ifCLASSINFOpdf
\else
\fi
\begin{document}
\title{\mytitle}

% \author{Michael~Shell,~\IEEEmembership{Member,~IEEE,}
%         John~Doe,~\IEEEmembership{Fellow,~OSA,}
%         and~Jane~Doe,~\IEEEmembership{Life~Fellow,~IEEE}% <-this % stops a space
% \thanks{M. Shell was with the Department
% of Electrical and Computer Engineering, Georgia Institute of Technology, Atlanta,
% GA, 30332 USA e-mail: (see http://www.michaelshell.org/contact.html).}% <-this % stops a space
% \thanks{J. Doe and J. Doe are with Anonymous University.}% <-this % stops a space
% \thanks{Manuscript received April 19, 2005; revised August 26, 2015.}}
\author{Zhenqiang~Ying\textsuperscript{\orcidicon{\myorcid}},%~\IEEEmembership{Student Member,~IEEE,}
and~Alan~C.~Bovik,~\IEEEmembership{Fellow,~IEEE}

\IEEEcompsocitemizethanks{\IEEEcompsocthanksitem 
The authors are with the Laboratory for Image and Video Engineering
Department of Electrical and Computer Engineering, 
The University of Texas at Austin,
Austin, TX 78712 USA.\protect\\
E-mail: zqying@utexas.edu; bovik@ece.utexas.edu%see http://www.michaelshell.org/contact.html
% \IEEEcompsocthanksitem J. Doe and J. Doe are with Anonymous University.
}% <-this % stops an unwanted space
\thanks{Manuscript received April XX, 20XX; revised August XX, 20XX.}}


\markboth{Journal of \LaTeX\ Class Files,~Vol.~14, No.~8, August~2015}%
{Shell \MakeLowercase{\textit{et al.}}: Bare Demo of IEEEtran.cls for IEEE Journals}



\maketitle

\begin{abstract}
% The abstract goes here.
Presenting context images to a viewer's peripheral vision 
is one of the most effective techniques to enhance immersive visual experiences.
% However, , because of the difficulty in preparing the context-image. 
However, most images only present a narrow view,
since the field-of-view~(FoV) of standard cameras is small.
% , which is one of the main reasons that contextual information is not as useful as it should be
% for object detection. 
To overcome this limitation, 
% In this paper, 
we propose a deep learning approach that learns to predict a 180-degree panoramic image from a narrow-view image.
Specifically,
we design a foveated framework that applies different strategies on near-periphery and mid-periphery regions.
Two networks are trained separately, and then are employed jointly 
to sequentially perform narrow-to-90\degree generation 
and 90\degree-to-180\degree generation.
The generated outputs are then fused with their aligned inputs to produce 
expanded equirectangular images for viewing.
% Then we design a foveated framework 
% First, two networks are trained separately to generate context images of near periphery and far periphery respectively.
% Then, 
% We design a two-stage foveated framework to meet the system recommendations.
% After that, we proposed a deep-learning method based on our framework. 
% Finally, we show experimental results and 
% Experiment results show that deep approaches to single-view-to-panoramic generation are both feasible and promising.
Our experimental results show that single-view-to-panoramic image generation using deep learning 
is both feasible and promising.
\end{abstract}


% \fi
\IEEEpeerreviewmaketitle

% \begin{multicols}{2}
% \vspace*{\textheight}
\input{\mymain}
% \columnbreak
% \lipsum[1-2]
% \end{multicols}

% or


% \appendices

% % you can choose not to have a title for an appendix
% % if you want by leaving the argument blank
% \section{}

% % use section* for acknowledgment
% \section*{Acknowledgment}


\ifCLASSOPTIONcaptionsoff
  \newpage
\fi



%\bibliographystyle{IEEEtran}


% \end{thebibliography}
% \bibliographystyle{IEEEtran}
% \bibliography{main.bib}

% \printbibliography
% \myIEEEbiography
% IEEEtran
\bibliographystyle{IEEEbib}
\bibliography{main.bib}

\begin{thebibliography}{10}

\bibitem{sabini2018painting}
Mark Sabini and Gili Rusak,
\newblock ``Painting outside the box: Image outpainting with gans,''
\newblock {\em arXiv preprint arXiv:1808.08483}, 2018.

\bibitem{jones2013illumiroom}
Brett~R Jones, Hrvoje Benko, Eyal Ofek, and Andrew~D Wilson,
\newblock ``Illumiroom: peripheral projected illusions for interactive
  experiences,''
\newblock in {\em Proceedings of the SIGCHI Conference on Human Factors in
  Computing Systems}. ACM, 2013, pp. 869--878.

\bibitem{turban2017extrafoveal}
Laura Turban, Fabrice Urban, and Philippe Guillotel,
\newblock ``Extrafoveal video extension for an immersive viewing experience,''
\newblock {\em IEEE transactions on visualization and computer graphics}, vol.
  23, no. 5, pp. 1520--1533, 2017.

\bibitem{avraham2011ultrawide}
Tamar Avraham and Yoav~Y Schechner,
\newblock ``Ultrawide foveated video extrapolation,''
\newblock {\em IEEE Journal of Selected Topics in Signal Processing}, vol. 5,
  no. 2, pp. 321--334, 2011.

\bibitem{aides2011multiscale}
Amit Aides, Tamar Avraham, and Yoav~Y Schechner,
\newblock ``Multiscale ultrawide foveated video extrapolation,''
\newblock in {\em Computational Photography (ICCP), 2011 IEEE International
  Conference on}. IEEE, 2011, pp. 1--8.

\bibitem{kimura2018extvision}
Naoki Kimura and Jun Rekimoto,
\newblock ``Extvision: Augmentation of visual experiences with generation of
  context images for a peripheral vision using deep neural network,''
\newblock in {\em Proceedings of the 2018 CHI Conference on Human Factors in
  Computing Systems}. ACM, 2018, p. 427.

\bibitem{patney2016foveated}
Anjul Patney, Marco Salvi, Joohwan Kim, Anton Kaplanyan, Chris Wyman, Nir
  Benty, David Luebke, and Aaron Lefohn,
\newblock ``Towards foveated rendering for gaze-tracked virtual reality,''
\newblock {\em ACM Transactions on Graphics (TOG)}, vol. 35, no. 6, pp. 179,
  2016.

\bibitem{guenter2012foveated}
Brian Guenter, Mark Finch, Steven Drucker, Desney Tan, and John Snyder,
\newblock ``Foveated 3d graphics,''
\newblock {\em ACM Transactions on Graphics (TOG)}, vol. 31, no. 6, pp. 164,
  2012.

\bibitem{thibos1987retinal}
LN~Thibos, FE~Cheney, and DJ~Walsh,
\newblock ``Retinal limits to the detection and resolution of gratings,''
\newblock {\em JOSA A}, vol. 4, no. 8, pp. 1524--1529, 1987.

\bibitem{lettvin1976seeing}
Jerome~Y Lettvin,
\newblock ``On seeing sidelong,''
\newblock {\em The Sciences}, vol. 16, no. 4, pp. 10--20, 1976.

\bibitem{rosenholtz2012rethinking}
Ruth Rosenholtz, Jie Huang, and Krista~A Ehinger,
\newblock ``Rethinking the role of top-down attention in vision: Effects
  attributable to a lossy representation in peripheral vision,''
\newblock {\em Frontiers in psychology}, vol. 3, pp. 13, 2012.

\bibitem{fridman2017sideeye}
Lex Fridman, Benedikt Jenik, Shaiyan Keshvari, Bryan Reimer, Christoph
  Zetzsche, and Ruth Rosenholtz,
\newblock ``Sideeye: A generative neural network based simulator of human
  peripheral vision,''
\newblock {\em arXiv preprint arXiv:1706.04568}, 2017.

\bibitem{fridman2017fast}
Lex Fridman, Benedikt Jenik, Shaiyan Keshvari, Bryan Riemer, Christoph
  Zetzsche, and Ruth Rosenholtz,
\newblock {\em A Fast Foveated Fully Convolutional Network Model for Human
  Peripheral Vision},
\newblock Verlag nicht ermittelbar, 2017.

\bibitem{weffers2011immersive}
A~Weffers-Albu, S~De~Waele, W~Hoogenstraaten, and C~Kwisthout,
\newblock ``Immersive tv viewing with advanced ambilight,''
\newblock in {\em Consumer Electronics (ICCE), 2011 IEEE International
  Conference on}. IEEE, 2011, pp. 753--754.

\bibitem{AmbiLux}
``Philips ambilight vs philips ambilux – what’s the difference?,''
  \url{https://www.whathifi.com/promoted/philips-ambilight-vs-philips-ambilux-whats-difference},
\newblock Accessed: 2019-01-12.

\bibitem{novy2013computational}
Daniel~Edward Novy,
\newblock {\em Computational immersive displays},
\newblock Ph.D. thesis, Massachusetts Institute of Technology, 2013.

\bibitem{xiao2016augmenting}
Robert Xiao and Hrvoje Benko,
\newblock ``Augmenting the field-of-view of head-mounted displays with sparse
  peripheral displays,''
\newblock in {\em Proceedings of the 2016 CHI Conference on Human Factors in
  Computing Systems}. ACM, 2016, pp. 1221--1232.

\bibitem{lubos2016ambiculus}
Paul Lubos, Gerd Bruder, Oscar Ariza, and Frank Steinicke,
\newblock ``Ambiculus: Led-based low-resolution peripheral display extension
  for immersive head-mounted displays,''
\newblock in {\em Proceedings of the 2016 Virtual Reality International
  Conference}. ACM, 2016, p.~13.

\bibitem{hashemian2018investigating}
Abraham~M Hashemian, Alexandra Kitson, Thinh Nquyen-Vo, Hrvoje Benko, Wolfgang
  Stuerzlinger, and Bernhard~E Riecke,
\newblock ``Investigating a sparse peripheral display in a head-mounted display
  for vr locomotion,''
\newblock in {\em 2018 IEEE Conference on Virtual Reality and 3D User
  Interfaces (VR)}. IEEE, 2018, pp. 571--572.

\bibitem{kimura2018using}
Naoki Kimura, Michinari Kono, and Jun Rekimoto,
\newblock ``Using deep-neural-network to extend videos for head-mounted display
  experiences,''
\newblock in {\em Proceedings of the 24th ACM Symposium on Virtual Reality
  Software and Technology}. ACM, 2018, p. 128.

\bibitem{cruz1992cave}
Carolina Cruz-Neira, Daniel~J Sandin, Thomas~A DeFanti, Robert~V Kenyon, and
  John~C Hart,
\newblock ``The cave: audio visual experience automatic virtual environment,''
\newblock {\em Communications of the ACM}, vol. 35, no. 6, pp. 64--73, 1992.

\bibitem{MITInfinity}
MIT Object Based~Media Group,
\newblock ``Infinity-by-nine,'' \url{http://obm.media.mit.edu/},
\newblock Accessed: 2019-01-12.

\bibitem{mills2011surround}
Peter Mills, Alia Sheikh, Graham Thomas, and Paul Debenham,
\newblock ``Surround video,''
\newblock 2011, p. 55–63.

\bibitem{jones2014roomalive}
Brett Jones, Rajinder Sodhi, Michael Murdock, Ravish Mehra, Hrvoje Benko,
  Andrew Wilson, Eyal Ofek, Blair MacIntyre, Nikunj Raghuvanshi, and Lior
  Shapira,
\newblock ``Roomalive: magical experiences enabled by scalable, adaptive
  projector-camera units,''
\newblock in {\em Proceedings of the 27th annual ACM symposium on User
  interface software and technology}. ACM, 2014, pp. 637--644.

\bibitem{Ariana}
Razer,
\newblock ``Project ariana,'' \url{http://www.razerzone.com/project-ariana},
\newblock Accessed: 2019-01-12.

\bibitem{pathak2016context}
Deepak Pathak, Philipp Krahenbuhl, Jeff Donahue, Trevor Darrell, and Alexei~A
  Efros,
\newblock ``Context encoders: Feature learning by inpainting,''
\newblock in {\em Proceedings of the IEEE Conference on Computer Vision and
  Pattern Recognition}, 2016, pp. 2536--2544.

\bibitem{yu2018generative}
Jiahui Yu, Zhe Lin, Jimei Yang, Xiaohui Shen, Xin Lu, and Thomas~S Huang,
\newblock ``Generative image inpainting with contextual attention,''
\newblock {\em arXiv preprint}, 2018.

\bibitem{zhang2013framebreak}
Yinda Zhang, Jianxiong Xiao, James Hays, and Ping Tan,
\newblock ``Framebreak: Dramatic image extrapolation by guided shift-maps,''
\newblock in {\em Proceedings of the IEEE Conference on Computer Vision and
  Pattern Recognition}, 2013, pp. 1171--1178.

\bibitem{xia2016novel}
Sifeng Xia, Shuai Yang, Jiaying Liu, and Zongming Guo,
\newblock ``Novel self-portrait enhancement via multi-photo fusing,''
\newblock in {\em Signal and Information Processing Association Annual Summit
  and Conference (APSIPA), 2016 Asia-Pacific}. IEEE, 2016, pp. 1--4.

\bibitem{shan2014uncrop}
Qi~Shan, Brian Curless, Yasutaka Furukawa, Carlos Hernandez, and Steven~M
  Seitz,
\newblock ``Photo uncrop,''
\newblock in {\em European Conference on Computer Vision}. Springer, 2014, pp.
  16--31.

\bibitem{sampetoding2018automatic}
Juan Leegard~Ranteallo Sampetoding, Bagus Satriyawibowo, Rini Wongso, and
  Ferdinand~Ariandy Luwinda,
\newblock ``Automatic field-of-view expansion using deep features and image
  stitching,''
\newblock {\em Procedia Computer Science}, vol. 135, pp. 657--662, 2018.

\bibitem{wang2014biggerpicture}
Miao Wang, Yukun Lai, Yuan Liang, Ralph~Robert Martin, and Shi-Min Hu,
\newblock ``Biggerpicture: data-driven image extrapolation using graph
  matching,''
\newblock {\em ACM Transactions on Graphics}, vol. 33, no. 6, 2014.

\bibitem{ashikhmin2001synthesizing}
Michael Ashikhmin,
\newblock ``Synthesizing natural textures.,''
\newblock {\em SI3D}, vol. 1, pp. 217--226, 2001.

\bibitem{ballester2000filling}
Coloma Ballester, Marcelo Bertalmio, Vicent Caselles, Guillermo Sapiro, and
  Joan Verdera,
\newblock ``Filling-in by joint interpolation of vector fields and gray
  levels,''
\newblock 2000.

\bibitem{barnes2009patchmatch}
Connelly Barnes, Eli Shechtman, Adam Finkelstein, and Dan~B Goldman,
\newblock ``Patchmatch: A randomized correspondence algorithm for structural
  image editing,''
\newblock in {\em ACM Transactions on Graphics (ToG)}. ACM, 2009, vol.~28,
  p.~24.

\bibitem{iizuka2017globally}
Satoshi Iizuka, Edgar Simo-Serra, and Hiroshi Ishikawa,
\newblock ``Globally and locally consistent image completion,''
\newblock {\em ACM Transactions on Graphics (ToG)}, vol. 36, no. 4, pp. 107,
  2017.

\bibitem{liu2018image}
Guilin Liu, Fitsum~A Reda, Kevin~J Shih, Ting-Chun Wang, Andrew Tao, and Bryan
  Catanzaro,
\newblock ``Image inpainting for irregular holes using partial convolutions,''
\newblock in {\em Proceedings of the European Conference on Computer Vision
  (ECCV)}, 2018, pp. 85--100.

\bibitem{yu2018free}
Jiahui Yu, Zhe Lin, Jimei Yang, Xiaohui Shen, Xin Lu, and Thomas~S Huang,
\newblock ``Free-form image inpainting with gated convolution,''
\newblock {\em arXiv preprint arXiv:1806.03589}, 2018.

\bibitem{bertalmio2003simultaneous}
Marcelo Bertalmio, Luminita Vese, Guillermo Sapiro, and Stanley Osher,
\newblock ``Simultaneous structure and texture image inpainting,''
\newblock {\em IEEE transactions on image processing}, vol. 12, no. 8, pp.
  882--889, 2003.

\bibitem{yang2017high}
Chao Yang, Xin Lu, Zhe Lin, Eli Shechtman, Oliver Wang, and Hao Li,
\newblock ``High-resolution image inpainting using multi-scale neural patch
  synthesis,''
\newblock in {\em Proceedings of the IEEE Conference on Computer Vision and
  Pattern Recognition}, 2017, pp. 6721--6729.

\bibitem{Yan_2018_ECCV}
Zhaoyi Yan, Xiaoming Li, Mu~Li, Wangmeng Zuo, and Shiguang Shan,
\newblock ``Shift-net: Image inpainting via deep feature rearrangement,''
\newblock in {\em The European Conference on Computer Vision (ECCV)}, September
  2018.

\bibitem{Xiong_2019_CVPR}
Wei Xiong, Jiahui Yu, Zhe Lin, Jimei Yang, Xin Lu, Connelly Barnes, and Jiebo
  Luo,
\newblock ``Foreground-aware image inpainting,''
\newblock in {\em The IEEE Conference on Computer Vision and Pattern
  Recognition (CVPR)}, June 2019.

\bibitem{allen1999screen}
Ioan Allen,
\newblock ``Screen size: The impact on picture and sound,''
\newblock {\em SMPTE Journal}, vol. 108, no. 5, pp. 284--289, 1999.

\bibitem{jiang2015salicon}
Ming Jiang, Shengsheng Huang, Juanyong Duan, and Qi~Zhao,
\newblock ``Salicon: Saliency in context,''
\newblock in {\em Proceedings of the IEEE conference on computer vision and
  pattern recognition}, 2015, pp. 1072--1080.

\bibitem{pix2pix2017}
Phillip Isola, Jun-Yan Zhu, Tinghui Zhou, and Alexei~A Efros,
\newblock ``Image-to-image translation with conditional adversarial networks,''
\newblock in {\em Proceedings of the IEEE conference on computer vision and
  pattern recognition}, 2017, pp. 1125--1134.

\bibitem{ronneberger2015u}
Olaf Ronneberger, Philipp Fischer, and Thomas Brox,
\newblock ``U-net: Convolutional networks for biomedical image segmentation,''
\newblock in {\em International Conference on Medical image computing and
  computer-assisted intervention}. Springer, 2015, pp. 234--241.

\bibitem{he2016deep}
Kaiming He, Xiangyu Zhang, Shaoqing Ren, and Jian Sun,
\newblock ``Deep residual learning for image recognition,''
\newblock in {\em Proceedings of the IEEE conference on computer vision and
  pattern recognition}, 2016, pp. 770--778.

\bibitem{mao2017least}
Xudong Mao, Qing Li, Haoran Xie, Raymond~YK Lau, Zhen Wang, and Stephen
  Paul~Smolley,
\newblock ``Least squares generative adversarial networks,''
\newblock in {\em Proceedings of the IEEE International Conference on Computer
  Vision}, 2017, pp. 2794--2802.

\bibitem{larsen2015autoencoding}
Anders Boesen~Lindbo Larsen, S{\o}ren~Kaae S{\o}nderby, Hugo Larochelle, and
  Ole Winther,
\newblock ``Autoencoding beyond pixels using a learned similarity metric,''
\newblock {\em arXiv preprint arXiv:1512.09300}, 2015.

\bibitem{arjovsky2017wasserstein}
Martin Arjovsky, Soumith Chintala, and L{\'e}on Bottou,
\newblock ``Wasserstein gan,''
\newblock {\em arXiv preprint arXiv:1701.07875}, 2017.

\bibitem{perez2003poisson}
Patrick P{\'e}rez, Michel Gangnet, and Andrew Blake,
\newblock ``Poisson image editing,''
\newblock {\em ACM Transactions on graphics (TOG)}, vol. 22, no. 3, pp.
  313--318, 2003.

\bibitem{xiao2012sun360}
Jianxiong Xiao, Krista~A Ehinger, Aude Oliva, and Antonio Torralba,
\newblock ``Recognizing scene viewpoint using panoramic place representation,''
\newblock in {\em Computer Vision and Pattern Recognition (CVPR), 2012 IEEE
  Conference on}. IEEE, 2012, pp. 2695--2702.

\bibitem{series2012methodology}
BT~Series,
\newblock ``Methodology for the subjective assessment of the quality of
  television pictures,''
\newblock 2012.

\bibitem{tyler1987analysis}
Christopher~W Tyler,
\newblock ``Analysis of visual modulation sensitivity. iii. meridional
  variations in peripheral flicker sensitivity,''
\newblock {\em JOSA A}, vol. 4, no. 8, pp. 1612--1619, 1987.

\end{thebibliography}

\end{document}